# Benchmarking for Metaheuristic Black-Box Optimization: Perspectives and Open Challenges

Ramses Sala
Department of Mechanical and Process Engineering
Technische Universität Kaiserslautern
Kaiserlautern, Germany
sala@rhrk.uni-kl.de

Ralf Müller
Department of Mechanical and Process Engineering
Technische Universität Kaiserslautern
Kaiserlautern, Germany
ram@rhrk.uni-kl.de

*Abstract*—Research on new optimization algorithms is often funded based on the motivation that such algorithms might improve the capabilities to deal with real-world and industrially relevant optimization challenges. Besides a huge variety of different evolutionary and metaheuristic optimization algorithms, also a large number of test problems and benchmark suites have been developed and used for comparative assessments of algorithms, in the context of global, continuous, and black-box optimization. For many of the commonly used synthetic benchmark problems or artificial fitness landscapes, there are however, no methods available, to relate the resulting algorithm performance assessments to technologically relevant real-world optimization problems, or vice versa. Also, from a theoretical perspective, many of the commonly used benchmark problems and approaches have little to no generalization value. Based on a mini-review of publications with critical comments, advice, and new approaches, this communication aims to give a constructive perspective on several open challenges and prospective research directions related to systematic and generalizable benchmarking for black-box optimization.

*Keywords— Evolutionary Algorithms, Metaheuristics, Black-Box Optimization, Benchmark, fitness landscape analysis*

## I. Introduction and Motivation

Aspects of mathematical optimization can sometimes be of practical and technological value. This is demonstrated by examples ranging from early applications such as the optimization of wine barrels by Johannes Kepler [1] as well as modern areas of engineering development [2]–[5] and other areas of technology [6]. Many technologically relevant simulation-based optimization problems are characterized by: a high number of design variables, non-linear multi-modal objective and constraint functions without gradient information. In an industrial context, there are many of such optimization problems and only limited resources to address them all. Therefore, metaheuristic optimization methods are often used as a compromise between optimization effort and accuracy (w.r.t. global optimality). Although a large variety of different optimizations algorithms (such as Evolutionary Strategies [7], [8], Genetic Algorithms [9], Particle Swarm Optimization [10], Differential Evolution [11], and many variants [12]–[14]) have been developed, it is still surprisingly difficult to identify efficient algorithms for a new instance of technologically relevant black-box optimization problem [15], [16].

Theoretical considerations showed that the quest for universally efficient search, learning, or optimization algorithms is futile [17]–[19]. The remaining open challenge is, therefore, to identify which algorithms perform well on particular problem instances or types [20]. This is sometimes also called the "per instance" or "per set" algorithm selection problem [16]. In the absence of theoretical methods to determine heuristic optimization algorithm performance on specific non-trivial problems, empirical algorithm performance analysis seems the last resort [21]. The history of physics and engineering demonstrated that a combination of theoretical and practical empirical approaches can be successful to establish quantitative models of complex phenomena and technological products. Several works in the field of Evolutionary and Meta-heuristic optimization, however, identified an increasing gap between theory and practice in the optimization communities [22], [23]. Also other works in the literature [19], [24]–[27] identified common shortcomings and misunderstandings in the theoretical, as well as practical aspects of optimization, and optimization research.

The aim of this communication is to formulate and highlight important open challenges, and prospective research directions, based on critical remarks, recommendations and work in the available literature, in order to improve the scientific and practical relevance of benchmarking for metaheuristic black-box optimization and continuous real-parameter optimization. Besides summarizing the relevant sections and statements from critical papers in the past, this work also aims to find and highlight the rare and relatively isolated research works-in the literature which did attempt to follow some of the advised research directions in order to set steps towards scientific generalizable optimization algorithm analysis, and industrially relevant benchmarks.

## II. The NFL theorems and their implications on benchmarking for metaheuristic optimization

The No Free Lunch (NFL) theorems and the Law for Generalization Performance for search, optimization, and learning [17]–[19] imply that if an algorithm performs well on one set of problems it must perform correspondingly poor on all other problems. Alternatively, for the setting of optimization: the average performance of all non-resampling algorithms is identical when averaged over all possible problems.

This research was partly funded by the DFG, (German Research Foundation) 252408385 – IRTG 2057

The NFL theorems invoked many discussions about their implications on competitive testing of optimization algorithms, and on their validity for real-world problems. Several works showed that for some problem classes, there are free "appetizers" [28] or "leftovers" [29] available. The work in [30] showed that on many subsets of all possible optimization problems, the assumptions of the NFL theorems are not fulfilled. It should also be noted that those subsets of problems are characterized exactly by the sort of properties that "real world" optimization problems might possess [31]. Also, other research works identified large general sub-sets of optimization problems where the NFL theorems do not hold [32], [33] (see also the review in [34]). These "Free Meal" (FM) studies seemed to have justified to continue the search for universal general-purpose metaheuristic algorithms that perform well on large sub-sets of optimization problems, in almost the same way as before publication of the NFL theorems.

Instead of ignoring or circumventing the implications of the NFL theorems, some works recognized the constructive implications of the NFL-theorems, for practical applications and scientific algorithm performance assessments. The theorems and their implications emphasize the limitations of generalizing optimization benchmark results to other problems. Although it has sometimes has been argued that the assumptions of the NFL theorems are irrelevant for practical applications, because real-world problems are likely to possess some exploitable structure, the theorems also imply that if the knowledge of the exploitable problem structure is not incorporated into a particular algorithm, no formal assurances exist that the algorithm will be effective [19]. From a practical point of view, the remaining challenge for Metaheuristic optimization is to match specific optimization algorithms with specific problems or classes of problems for which those algorithms are well suited and perform relatively effectively. The suitability or efficiency depends on the exploitation of the problem structure. Which leads to interesting open challenges of practical and theoretical relevance: *How can we construct algorithms that use the available knowledge or assumptions on the problem structure?* And of more fundamental importance: *How can we determine, describe, and quantify the problem structure?*

The implications of [17], [19] caused the need to set steps in a new research direction, which should: *"focus attention on tailoring algorithms and representations to particular problem classes by exploiting domain knowledge"* [35]. The article also indirectly introduced a new way for optimization algorithm development and competitions, by proposing the concept of out-of-sample testing to evaluate algorithm performance on particular well-specified problem classes. *"The idea would be to refine algorithms using any small, randomly chosen subset of problems from this class, but to compare performance only against different randomly selected problem instances from the same class"* [35]. Although similar concepts are now standard practice in the field of machine learning, these ideas, have received remarkably little attention within the community of black-box optimization (with [36] as a special exception), and the related communities on real-parameter optimization and continuous optimization using metaheuristics.

In [37] the importance of the assumptions one can make about the sub-set or class of optimization problem instances was highlighted. It was also remarked that: *"Unfortunately, general theories about the nature of likely subsets are lacking"* [37], and that such theories are needed to identify efficient algorithms for practical applications. The review in [34] concluded that: to evaluate optimization algorithms, benchmarking alone is not enough; it must be used in combination with clear underlying assumptions on the distribution of the targeted problem subset. *"The benchmark functions must be representative of the problems and there must be some smoothness, in the sense that being good at a problem means that the algorithm is likely to be good at similar problems"* [34]. For many real-world black-box optimization problems it is not even clear yet what "similarity" means. The following section summarizes critical comments, advice, and results from previous works related to generalization value in optimization algorithm benchmarking.

### III. BENCHMARKING AND SCIENTIFIC ALGORITHM PERFORMANCE ANALYSIS

*"If the conditions of No Free Lunch are violated then some algorithm can have best expected performance, but this does not help us to find it, or to say anything about it."* [38]. When the NFL theorems do not apply for a given subset of problems, the quest to identify dominating algorithms seems justified [33], [39]. For non-convex multi-modal optimization problems, theoretical algorithm performance analysis methods are very limited. Therefore "empirical" comparative assessments, using numerical experiments on test problems or benchmark functions, are commonly used [21]. Besides a huge variety of different meta-heuristic optimization algorithms, also a large number of exotic benchmark problems and collections have been developed, and used in various competitions [40]–[49] Many of these benchmark problems are artificial-landscape-based problems, which are difficult to relate to real-world problem instances and vice versa [22], [50], [51].In many cases they are even difficult to relate to other synthetic problem instances. Also, from a theoretical perspective, commonly used benchmark problems have been criticized for regularities and simplicity [52], [53]. Even more fundamental issues are the substantial lack of generalization value [27] and the lack of systematism in state-of-the-art benchmark-based comparative assessments [24], [54].

In the context of optimization algorithm benchmarking, generalization value refers to whether an algorithm performance assessment has the scope to estimate algorithm performance for unseen problem instances of the same class or a specified set of problems. To investigate generalization thus involves the statistical quantification of expected performance of the algorithms on a specific domain of problems, as well as specifying and quantifying the bounds or the "size" of the domain of representativeness of the experimental benchmark set up. Common shortcomings of benchmark-based results can be illustrated by the following two scenarios and accompanying questions:

*1 If a statistically significant comparative test indicates that algorithm A performs better than algorithm B with a probability of P, on some benchmark function $f_{123}$. What general value do such results have? What is the set or domain of problems for which the presented results are expected to have generalization value?*

*2 Even in an idealized framework where all metaheuristic algorithms published till today would be available, and test*

*results on the union of the currently commonly used benchmark functions, according to respected reporting standards (such as [55]) are available. What generalization value would new benchmark results involving new algorithms or new test problems accomplish? Do the benchmark results contribute any information that can be used to make a reasonable statement about expected algorithm performance on an unseen (not explicitly tested) optimization problem instance of a specific problem set?*

Although there are benchmark studies that do contribute generalizable results, they are unfortunately still a minority. In [24] the critical observation was made that some have confused science with development. In the empirical sciences, experiments are designed to answer a question or thesis, or in the more quantitative empirical sciences such as physics and engineering to calibrate or validate a model (see also [56]). The value or impact of experimental results is not only determined by its reproducibility and statistical significance, but also to the relevance and scope of the model or research question. If the question or model has no general scientific relevance, then the result of the experiment is likely to have little value and does not really need to be published in a scientific venue. There exists, however another category of experiments in the field of engineering development, which have little generalization value, but are relatively common and suit their purpose. These are the experiments which at the end of a product development process, test the safety or quality of a specific product under specific circumstances. In engineering, such experiments are however rarely published in scientific venues, while in the optimization community, similarly specific results with little general relevance are quite commonly published.

The work of [24] called for a more systematic and scientific approach for testing Metaheuristics. The work argued for a change in direction of the heuristic optimization community to develop and apply methods for controlled experimentation similar to approaches used in the empirical sciences, instead of competitive testing. As an outlook and direction for future work, the article suggested the development of methods to construct problems where parameters that may influence performance can be controlled. *"The problems are not only likely to be atypical but deliberately so, in order to isolate the effect of various characteristics. Admittedly, the choice of which factors to control for is far from trivial and may demand considerable insight as well as trial and error. But it is a problem that creative scientists deal with successfully in other disciplines, whereas the task of choosing representative benchmark problems seems to confound all efforts. Furthermore, it is a problem that algorithmists ought to struggle with because it goes to the heart of what empirical science is all about."*[24].

In order to make meaningful optimization algorithm performance assessments with scientific relevance, more systematic and generalizable approaches to optimization algorithm benchmarking are required. Only very few works heading in this direction can be found in the literature. In [57] rational functions with prescribed global and local minimizers were presented. The resulting functions could be adjusted to match various difficulties, such as high multi-modality, ill-conditioned Hessian matrices and narrow deep holes. A general-purpose tuneable landscape generator was presented in [58]. The resulting benchmark functions are parameterized in a way to have an intuitive effect in terms of geometric features of the landscape. A study where benchmark problem test instances were regarded as drawn from a population with particular characteristics, in order to analyze optimization algorithms for a specific class of problems was presented in [59]. The landscape generator from [58] was combined the statistical approach from [59] in [27], where a new approach was proposed to create generalizable benchmarking results. The presented approach enabled the generation of real-world data-based natural problem classes, from which new test problem instances could be generated for statistical analysis of optimization algorithm performance. In [60] the fitness landscapes of clustering problems were analyzed and used for optimization algorithm benchmarking. Although this an excellent example of generalizable benchmarking for clustering problems, the field would benefit from systematic approaches for other classes of problems. A benchmarking approach based on weighted composition of structured random fields was proposed in [54]. There presented function generator enables the systematic construction of test problems which are parameterized w.r.t. various function features such as dimension, multi-modality, noise, first and higher-order global sensitivity indices (which describe the effective dimension and variable interactions). The above mentioned parameterized benchmark problem approaches provide a description of the feature space, in which distance metrics could be defined, that could be of use to quantify specific problem classes. Compared to the conventional benchmark functions and suites, the mentioned works however received relatively little response from the community.

## IV. TECHNOLOGICALLY RELEVANT OPTIMIZATION BENCHMARKS

Because research on metaheuristic optimization algorithms is often motivated and funded in perspective of its potential technological and real-world relevance, it seems reasonable to assess the relevance of commonly used benchmarks for real-world problems. It has been noted that the 'gap' in technology transfer between the optimization algorithm development and engineering applications is "*partially due to the nonexistence of practical benchmark problems*" [61]. This observation was also confirmed in the statement made in [22] that there is a gap between the "toy" problems often used in theory and development of algorithms and the complexity of real-world problems. A perspective on possible reasons for the mismatch between *evolutionary computation* research and the number of actual real-world applications is provided in [25]. Where some of the possible causes were identified as: unrepresentative benchmark problems (small-scale silo problems); the focus on global optima (which are often computationally infeasible for real-world problems in an industrial setting); and the dislike of business applications in the research community [25]. In another paper [23] the same author stated two main reasons for the growing gap between theory and practice in evolutionary computing:

1   The growing complexity of real-world problems

2   The focus of the research community on issues that are secondary for real-world applications.

One of the suggested directions for further research related to the first point was: *"the development of artificial problems that better reflect real-world difficulties, which the research community can use to experience (and appreciate) for themselves what it really means to tackle a real-world problem"* [23]. Although other optimization communities such as

Operation Research and Engineering Optimization are more application-oriented, similar calls for more realistic and complex benchmark optimization problems have also been expressed by other authors [22], [61], [62]. For simulation-based benchmarks, several attempts have been made to initiate public repositories [63]–[66], but these initiatives did unfortunately not become very popular, therefore improved industrially relevant optimization benchmarks and repositories are needed [62].

Industrial simulation-based engineering design optimization problems often involve computationally expensive function evaluations. Paradoxically: the problems for which optimization performance matters the most because they are computationally expensive and in practice restricted to a limited function evaluation budget are also the problems for which it is often too expensive to benchmark algorithms, tune the optimization parameters, or develop specialized optimization methods [50]. The development and dissemination of new computationally expensive simulation-based benchmarks are therefore only of limited value. Besides developing and sharing technologically relevant benchmarks, also the function evaluation data based on pseudo or quasi-random sampling in the design space and the results of the corresponding simulations could be valuable. Such data could be used to construct response surfaces or surrogate models. Such surrogate models could serve as benchmark problems, or as input to analyze the simulation response characteristics. It should be noted, however, that the results of surrogate-based optimization algorithm performance assessments are not always representative for algorithm performance on the high-fidelity simulation-based objective and constraint functions [67]. This could be caused by the smoothing effect of the surrogate model and the limited resolution of the computationally expensive data of the training set.

Alternatively, to conventional data-based meta-models and surrogate models for optimization benchmarking, some new ideas related to feature-based surrogate models for benchmarking have been presented and investigated in [50], [51], [68]. The general idea behind these concepts is to construct computationally affordable representative benchmark problems, based on replacement or surrogate functions which capture the characteristics of the underlying high-fidelity simulation responses. These benchmarks are then used for algorithm selection and parameter tuning, after which the most promising approach is applied to the expensive real-world optimization problem. The approach in [51] was based on a quantitative simulation-response characterization, on various instances of a similar simulation-based optimization problem. A constraint satisfaction problem was formulated, in which the identified problem characteristics were imposed as constraints on a set of parameterized functions, in order to generate an optimization problem with similar characteristics. Independently a similar approach was presented in [68], with analytic replacement problems that preserve the functional characteristics of the original problem, targeting to benchmark various Multidisciplinary Design Optimization architectures. Research to extricate the benchmarking problem for computationally expensive problems is rare. Although the here mentioned approaches seem to have potential, the related case studies were rather specific. Therefore, further investigations or alternative ideas in this direction are needed.

In an industrial engineering setting, often many distinct instances of similar design problems need to be solved (e.g., in Automotive, Aircraft, Windmill engineering). Large development projects are composed of many optimization tasks, and only limited resources are available. For industrial design and engineering applications, the efficiency of algorithms in a limited function evaluation budget setting is therefore of particular importance, because it directly influences product and process performance. Due to the high-performance demands and recurring problem instances, algorithms, and parameter settings that are effective of specific problem classes are therefore important. Another issue related to industrially relevant optimization benchmarking is the question: *How representative are benchmark results of one optimization problem instance to the algorithm performance on another problem instance?* The answer is usually application and problem class dependent. This topic has been addressed for "facility layout planning" [69], and extensively for "online bin packing" problems in [70], [71]. However, for complex simulation-based problems, and "tagged" black-box optimization problem classes (either synthetic or real-world) only few works in the literature address or even touch the issues related to benchmark representativity and generalizability. This despite the topic's fundamental importance for industrial optimization.

V. THE ALGORITHM SELECTION PERSPECTIVE

Although optimization algorithm benchmarking is not necessarily performed for reasons related to algorithm selection. The general algorithm selection framework presented in [15] can provide a valuable perspective. The framework has four main components: the problem, algorithm, performance, and characteristic spaces (see also the review in [72] for further details). In a nutshell, optimization algorithm selection is about matching an optimization problem with an optimization algorithm that performs well on that problem. This involves two important factors: 1) problem characterization and classification, and 2) Algorithm performance assessment, and which will be discussed in the following sub-sections.

A. *Problem characterization and categorization*

Black-box optimization algorithm performance assessments are often based on collections of rather isolated benchmark problems [40]–[43], [45]–[47], [49]. Generally, optimization algorithm performance on such benchmark problem instances is rather difficult to relate to performance on other problems [27], [54]. Although the various approaches of problem characterization also called "fitness-landscape analysis" [73] or "Explanatory Landscape Analysis" (ELA) [74] could be important in the context of algorithm selection [16], the topic has received relatively little attention in the literature compared to the dissemination related to new algorithms [75]. Although interesting developments are presented in [74], [76]–[78], it was noted in [72] that most works on ELA, attempt to provide a single measure for optimization problem complexity, and that further research to complementary measures was needed. Therefore, a further important open question to address is:

*How to obtain, describe and quantify knowledge about features and characteristics of specific optimization problem classes and instances?*

Or in other words: How to effectively characterize and categorize optimization problems? A possible approach could be to combine several ELA criteria to establish a multidimensional description of the problem instance space in order to describe discriminate and relate various optimization problems and problem classes. Problem characterization and categorization seem important for simulation-based applications of technological relevance as well as for synthetic benchmark problems for analytical and scientific purposes. While synthetic benchmark problems could be constructed with targeted characteristics, for true black-box optimization problems characterization requires sampling-based approaches to determine or estimate the problem characteristics. Methods and concepts from the field of global sensitivity analysis could be used to describe and characterize some of the features of optimization problems [51], [79], [80]. For computationally expensive black-box problems surrogate models could be used as an intermediate step for the characterization. The quality and accuracy of the characterization are then dependent on the fidelity of the surrogate model, which depends on the sample size of black-box function evaluations. For computationally expensive problems such characterizations might still be costly, but when many distinct instances of a particular problem class need to be evaluated, (which is regularly the case in industrial settings) it might be worth the investment.

## B. Optimization algorithm performance assessment

According to [20] optimization algorithm performance can be seen as the result between an (abstract) inner product of two vectors. Where the first vector contains all the details of how the search algorithm operates, and the second contains all the details of the problem. At present, no specific methods to express either problem or algorithm properties in a vector notation suitable convenient performance calculation are available. In the absence of such methods, the most reasonable alternative seems to use systematic generalizable empirical benchmarks to determine algorithm performance. This leads to the question:

*How to exploit knowledge or assumptions on particular properties of problem sub-sets or classes, in the selection or development of effective optimization strategies?*

One possible scenario is empirical reverse engineering: Given a problem instance or problem class, with a sufficient description of its characteristics (see the open challenges in the previous sub-section). One looks in a collection with generalizable benchmarks, for results obtained on problems with similar characteristics. Alternatively, one could try to create computationally affordable representative benchmark problems using one of the ideas presented in (sections III and IV). Based on the benchmark results of either way one could then select a suitable optimization algorithm. A conceptually similar approach to the sketched idea for algorithm selection has been proposed and applied to the graph coloring problems in [81]. Taking this idea one step further to algorithm development, one could use genetic programming or another an optimization algorithm that is parameterized w.r.t. its operators and parameters to optimize the algorithm performance on a specific class of problems, with specific characteristics.

There are two essential differences in the approach sketched in the previous paragraph and the conventional comparative assessment approaches for algorithm selection with commonly used benchmarks:

1 *Feature-based benchmarking*: Relevant test problem instances are selected based on similarity with the targeted problem class characteristics (or its assumed features). Instead of treating the problem as a completely untagged black-box without any assumptions about its characteristics.

2 *Out-of-sample benchmarking*: The benchmark-based performance assessment results and algorithm selection are obtained using one or more problem instances which are distinct (but similar in terms of features) from the targeted problem. Rather than testing on a specific single problem or a set of isolated disconnected problems.

As long as no relations or a measure of distance between the various optimization problems is formulated, no generalization or relevance of benchmark results to other problems can be established or justified. Generalization in optimization requires a suitable representation of the characteristics of a problem class. A few newly developed approaches targeted generalizable algorithm performance assessments for sub-sets or classes of problems with specified characteristics [14], [27], [57], [58], [60]. Further development and use of such benchmark and assessment approaches could enable to select, develop, and tune optimization algorithms for specific problem characteristics. Feature-based benchmark approaches can also contribute to the development, training, and assessment of self-adaptive algorithms, and hyper-heuristics [82]. Although out-of-sample testing on unseen problem instances is required to test for generalization [35], [83], this is rarely applied in the context of global, continuous, and black-box optimization.

Although some works have shown or argued that the NFL theorems [17], [19] do not hold on many problem classes of practical relevance [28]–[33], [39], the practical implication that optimization problems should be tailored to the problem characteristics remains a valuable insight. For a true black-box optimization problem strictly no generalization to and from other problems is practically possible. In practice, however, often some additional properties are known or assumed and can be "tagged" on black-box optimization problems or problem classes. In [84], several open challenges of practical and theoretical relevance to the optimization community were presented. Also, the description and quantification of problem-specific knowledge were highlighted. *"we have to seek balance between specialty and generality, between algorithm simplicity and problem complexity, and between problem-specific knowledge and capability of handling black-box optimization problems"* [84]. Combining this idea with those presented in [20]. One could argue that the challenge for research in heuristic optimization is not only to match or tailor effective algorithms for particular problem classes, but also to balance algorithm and problem domain specificity and generality. Where the NFL and FM theorems target large general domains of optimization problems which can be characterized as "disorganized complexity", many practitioners and algorithm developers in the optimization community are currently mainly exploring "toy" and benchmark problems of "organized simplicity" (specific problems of relatively short minimum problem description length). Similar as in complex systems theory [85], it is the middle region of optimization problems with "organized complexity" that spans the difficult open challenges of scientific and technological relevance.

## VI. SUMMARY OF OPEN CHALLENGES AND PROSPECTIVE RESEARCH DIRECTIONS

This communication highlights contents, critical comments, and constructive advice from over 80 references in order to identify and formulate important open challenges and prospective directions for future research and development related to benchmarking and algorithm selection for black-box optimization, and related areas of optimization.

For the setting of true black-box optimization, without any other assumptions on a given and uninvestigated problem the NFL theorems apply, and consequently no algorithm preference can be justified based on the priors. For some specific subclasses of optimization problems or 'tagged' black-box problems, the NFL theorems do not apply. This does, however, not automatically imply that comparative benchmark results of algorithms on some problem instances of such a class have any generalization value to other problems in the class. This applies to synthetic as well as real-world-based benchmark problems. Therefore, the scientific value and the practical relevance for real-world optimization problems, of many commonly used benchmark set-ups are contentious. The scientific or technological value of an empirical result is related to its reproducibility, statistical significance, and the size or scope of its generalization domain. In order to establish a non-trivial generalization domain of benchmark-based algorithm performance assessment, the targeted domains need to be specified. Furthermore, to test any assumed or estimated generalization value of benchmark results, out-of-sample testing or problem-class-based benchmarking is necessary. To enable knowledge transfer between synthetic problem-based benchmark results, and real-world optimization problems, methods to determine and describe problem features or characteristics are needed. The specification of the generalization domain for an optimization algorithm benchmark study, requires to balance the specificity and generality of the targeted scientific scope and practical value of the investigations.

The framework for algorithm selection provides valuable perspectives for optimization algorithm development and analysis. In this context two important open challenges and corresponding directions for future work are identified:

*1 How to obtain, describe, and quantify knowledge about the features and characteristics of specific optimization problem classes and instances?*

Related to the following directions for future work:

- Identification of various criteria to describe essential optimization problem features and characteristics.
- Development of new methods to quantitatively characterize optimization problems according to specific criteria (quantitative fitness-landscape analysis).
- Establishment of multidimensional feature space descriptions to discriminate and relate different optimization problem instances with respect to a "basis" of features and corresponding distance metrics.
- Development of a framework for the categorization or classification of optimization problems according to their characteristics.

*2 How to exploit knowledge or assumptions on particular properties of problem sub-sets or classes, in the selection or development of effective optimization strategies?*

Related the following directions for future work:

- Development and extension of benchmark methods that enable to obtain generalizable results, and systematic investigation of the influence of problem characteristics on algorithm performance behavior.
- Out-of-sample testing and benchmarking of optimization algorithms.
- Development of algorithm operators that exploit particular combinations of problem features.
- Reverse Engineering: Empirical algorithm testing on generalizable benchmark problems, to identify which algorithm operators or settings perform well on particular problem features.

Besides the previous points of general importance, other open challenges, and directions of technological and industrial importance are:

- Formulation, implementation, and dissemination of accessible optimization problems of industrial importance.
- Sharing data-sets containing the function evaluation history on expensive simulators using pseudo-random or quasi-random samples to serve as a basis for surrogate models or landscape characterization.
- Investigations on the similarity and variation of problem instances within classes of real-world optimization problems.
- Considerations about how specific or general problem classes should be defined to be of practical use.

A summarizing perspective is that the essential challenge of optimization algorithm benchmarking in the algorithm selection context is to achieve systematic generalizable matching of effective algorithms for particular problems or problem classes. The essential challenge in optimization algorithm benchmarking in the context of algorithm analysis and development, goes beyond matching algorithms and problems and aims to investigate and exploit the effect of algorithm operators on specific problem characteristics. In both settings, the characterization and classification of the benchmark problems and problem sets for the targeted generalization domain is essential to obtain results of scientific and practical value. For the setting of black-box optimization, generalizable benchmark results can only be obtained for specific classes of "tagged" black-box problems. Therefore, suitable criteria, and systematic approaches to "tag", characterize, describe, and relate black-box problem classes and problem instances are needed. Theoretical frameworks and empirical approaches that enable to bridge the extreme generality of the NFL theorems and FM studies on one side, and the extreme specificity (and lack of generalizability) of commonly used benchmark set-ups on the other side, are required to make progress of scientific and practical relevance.